\documentclass{article}
\usepackage{kr}

\usepackage{times}
\usepackage{soul}
\usepackage{url}
\usepackage[hidelinks]{hyperref}
\usepackage[utf8]{inputenc}
\usepackage[small]{caption}
\usepackage{graphicx}
\usepackage{amsmath}
\usepackage{amsthm}
\usepackage{booktabs}
\usepackage{algorithm}
\usepackage{algorithmic}
\title{Automated Interactive Domain-Specific Conversational Agents\\ that Understand Human Dialogs}

\author{%
Yankai Zeng$^1$\and
Abhiramon Rajasekharan$^1$\and
Parth Padalkar$^1$\\
Kinjal Basu$^2$\and
Joaqu\'{\i}n Arias$^3$\and
Gopal Gupta$^1$\\[.5em]
\affiliations
$^1$Depart of Computer Science, UT Dallas, USA\\
$^2$IBM T. J. Watson Research Center, NY, USA\\
$^3$CETINIA, Universidad Rey Juan Carlos, Madrid, Spain\\[.25em]
\emails
\{Yankai.Zeng, Abhiramon.Rajasekharan, Parth.Padalkar, Gopal.Gupta\}@utdallas.edu, 
Kinjal.Basu@ibm.com,
Joaquin.Arias@urjc.es
}

\usepackage[dvipsnames]{xcolor}
\definecolor{PrologPredicate}{RGB}{0,0,200}
\definecolor{PrologVar}      {RGB}{145,032,039}
\definecolor{PrologComment}  {RGB}{169,082,044}
\definecolor{PrologOther}    {rgb}{0.2,0.2,0.2}
\definecolor{PrologString}   {RGB}{070,120,200}
\usepackage{relsize}
\usepackage{listings} 

\lstdefinestyle{MyInline}
{
  basicstyle = \ttfamily\color{PrologOther},
  breaklines = true,
  breakatwhitespace=true,
  upquote = true,
}
\lstdefinestyle{MySCASP}
{
  keywords = {},
  upquote = true,
  basicstyle = \relsize{-.5}\ttfamily\color{PrologPredicate},
  moredelim = {*[s][\color{black!40!PrologPredicate}]{\#pred}{.}},
  moredelim = {*[s][\color{black!40!PrologPredicate}]{\#show}{.}},
  moredelim = {*[s][\color{black!40!PrologPredicate}]{\#hide}{.}},
  moredelim = {*[s][\color{PrologVar}]{(}{)}},
  moredelim = {*[s][\color{PrologString}]{'}{'}},
  moredelim = {*[s][\color{PrologOther}]{:-}{.}},
  commentstyle = \mdseries\color{PrologComment},
  morecomment=[l]\%,
}
\usepackage{todonotes}
\usepackage{enumitem}
\newcommand{\citealp}[1]{\citeauthor{#1} (\citeyear{#1})}

\begin{document}

\maketitle

\begin{abstract}
Achieving human-like communication with machines remains a classic, challenging topic in the field of Knowledge Representation and Reasoning and Natural Language Processing. While Large Language Models (LLMs) have shown promise in generating human-like sentences for tasks such as question answering, paragraph summarization, and translation, they rely on pattern-matching rather than a true understanding of the semantic meaning of a sentence. As a result, they may generate incorrect responses. To generate an assuredly correct response, one has to ``understand” the semantics of a sentence, so that the missing information can be further requested and the correct response computed. To achieve this ``understanding”, logic-based (commonsense) reasoning methods such as Answer Set Programming (ASP) are arguably needed. In this paper, we describe the AutoConcierge system that leverages LLMs and ASP to develop a conversational agent that can truly ``understand” human dialogs, at least in restricted domains. AutoConcierge is focused on a specific domain---advising users about
% Abhiramon: below means restaurants in a certain area?
restaurants in their local area based on their preferences. AutoConcierge will interactively understand a user’s utterances, identify the missing information in them, and request the user via a natural language sentence to provide it. Once AutoConcierge has determined that all the information has been received, it computes a restaurant recommendation based on the user-preferences it has acquired from the human user. AutoConcierge is based on our STAR framework developed earlier, which uses GPT-3 to convert human dialogs into predicates that capture the deep structure of the dialog’s sentence. These predicates are then input into the goal-directed s(CASP) ASP system for performing commonsense reasoning. To the best of our knowledge, AutoConcierge is the first automated conversational agent that can realistically converse like a human and provide help to humans based on truly understanding human utterances. 
%We also present a general framework for developing such domain-specific automated conversational agents.
%Achieving human-like communication with machines remains a classic and challenging topic in the field of Knowledge Representation and Natural Language Processing. While Large Language Models (LLMs) have shown promise in generating human-like sentences for tasks such as question answering, paragraph summarization, and translation, they rely on pattern-matching rather than a true understanding of the semantic meaning of the sentence. As a result, they lack the ability to reason deeply or ensure correctness. In contrast, logic-based reasoning tools such as Answer Set Programming (ASP) offer soundness and high explainability.

% Is it good to mention the STAR framework here?
%In this paper, we present a conversational agent that leverages both LLMs and logic-based reasoning to truly understand human language and behave like a human. Following the framework of \textbf{STAR} \cite{star}, we use LLMs to extract information from the text and turn it into predicates, and then add a reasoning module for deep understanding. Our agent is focused on providing restaurant recommendations but also demonstrates generalizability to other conversational topics.

%In addition, we propose a philosophy for designing task-based conversational agents, including \textbf{awareness}, \textbf{proactivity}, \textbf{correctness}, and \textbf{consistency}. We believe this approach represents the correct path toward achieving \textit{TRUE} artificial intelligence.
\end{abstract}

\section{Introduction}
Conversational agents are designed to understand dialogs and generate meaningful responses to communicate with humans. The recently popular ChatGPT, with its surprising performance and powerful conversational ability, 
% Abhiramon: can we rephrase below as 'show promise for using LLMs ... for the vexing task of conversational AI.'
brought \textit{Large Language Models} (LLMs) such as GPT-3\cite{gpt3}, PaLM \cite{palm}, and LLaMa \cite{llama} as the solution to the vexing problem of developing conversational AI systems. These LLMs work quite well in content generation tasks such as translation and creative writing, but their deficiency in fact-and-knowledge-oriented tasks is well-established by now \cite{chatgpt-critic}. These models themselves cannot tell whether the text they generate is based on facts or made-up stories, and they cannot always follow the given data and rules strictly and sometimes even modify the data at will. The reasoning that these LLMs appear to perform is also at a very shallow level. These are serious flaws that make the LLMs unsuitable for task-driven conversations such as providing correct information to a user.

The reason for the above issues is that LLMs generate text that is purely based on a pattern-matching mechanism. As a result, LLMs have absolutely no understanding of the meaning of sentences and thus lack \textit{awareness}  \cite{chatgpt-critic}.  In contrast, humans understand the meaning of sentences, then use their reasoning capabilities to draw further conclusions, check for consistency, or determine missing information from this meaning. Thus, to make the machine-generated response reliable and consistent, we need to follow a similar approach. We use LLMs to extract knowledge---represented as predicates---from sentences. Next, we use logic-based methods such as answer set programming (ASP)---that facilitates commonsense reasoning---for drawing further conclusions, checking the consistency of information, or determining missing information with the help of these predicates. Generation of the response given to the user also leverages LLMs by translating predicates computed by the ASP engine into natural language sentences. This approach to developing intelligent systems has been encapsulated in the 
% Abhiramon: This STAR system is from ICLP paper right? If so, the reference needs to be changed.
% Solved
STAR framework \cite{star} that we developed earlier. 
%In this paper we show how this framework can be used to develop an elaborate goal-oriented conversational agent. 

\begin{figure}[ht]
    \vspace{-0.2in}
    \centering
    \includegraphics[width=0.4\textwidth]{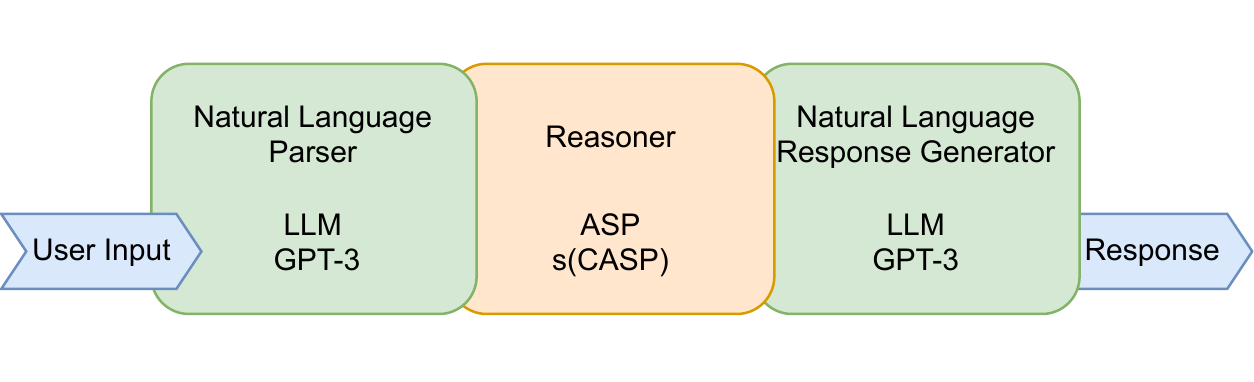}
    \caption{The AutoConcierge Architecture}
    \label{fig:design}
    \vspace{-0.15in} 
\end{figure}

Following the above insights, in this paper we report on developing an elaborate conversational agent that can understand human dialog and respond properly according to human intention. We narrow the domain of our conversational agent to give advice about finding restaurants in the user's vicinity.
% Abhiramon: If we plan to use the term 'area restaurants', we should maybe define it prior.
%area restaurants. 
We call the agent AutoConcierge. AutoConcierge is built with restaurant information around the University of Texas at Dallas, and is able to provide precise information based on user preferences. User preference is elicited by AutoConcierge via a natural language dialog with the human user. AutoConcierge will first ask the user a few questions to which the user responds in natural language. Once enough information is collected, AutoConcierge will find a restaurant in its knowledgebase that satisfies (most of) the user's requirements, and that does not violate any restriction imposed by the user. Users can also ask AutoConcierge for other possible recommendations, or even modify their requirements mid-conversation. All these abilities are achieved by \textit{reasoning} through the structured predicate expression captured from the user's input. These predicates are generated through the use of LLMs. Figure \ref{fig:design} shows the high-level architecture of our AutoConcierge system. Following our STAR framework developed earlier \cite{star}, we use GPT-3 to ``translate" the dialog from English to predicates, and then feed these predicates to the goal-directed s(CASP) ASP system. The s(CASP) system holds commonsense knowledge for making restaurant recommendations and computes a response as logical term(s) or predicate(s). The term(s) or predicate(s) are then converted into human-understandable natural language expression by GPT-3 as well. Detailed implementation of the AutoConcierge system is discussed in the rest of this paper.

To the best of our knowledge, AutoConcierge is the first conversational agent to communicate with humans based on truly understanding human utterances. It leverages both the LLM as well as ASP technology. Crucially, it relies on the s(CASP) goal-directed ASP system \cite{scasp}. The AutoConcierge system has several advantages over an approach that is purely based on LLMs:
\begin{itemize}
    \item AutoConcierge can check if the knowledge extracted from the user utterance is consistent and correct. This is because it possesses commonsense knowledge about the domain represented as ASP rules \cite{gupta-csr}.
    \item It gives reasonable advice based on the knowledge of the user's likes and dislikes, which LLM-only agents are not able to explicitly consider.
    \item It is capable of precisely justifying its recommendations. 
\end{itemize}

Due to a lack of objective criteria for evaluating the quality of responses produced by AutoConcierge, it is hard to compare different conversational agents. To address this problem, we also propose a list of criteria for evaluating task-driven conversational agents. These include \textit{proactivity}, \textit{economy}, \textit{explainability}, \textit{correctness},  \textit{consistency}, and \textit{efficiency}. These criteria are further discussed later.
%time-consumed should be part of efficiency
%We also provide a general method for developing task-driven automated conversational agents, similar to AutoConcierge, for other domains.

%The rest of the paper is organized as follows: 

\section{Background}
\subsection{Large Language Models}
Until recently, transformer-based deep learning models have been applied to NLP tasks by training and fine-tuning them on task-specific datasets \cite{pre_trained_transformers}. With the advent of Large Language Models, the paradigm changed to teaching a language model any arbitrary task using just a few demonstrations, called \textit{in-context learning}. Brown et al introduced an LLM called GPT-3 \cite{gpt3} containing approximately 175 billion parameters that has been trained on a massive corpus of filtered online text, on which the well-known ChatGPT is 
% Abhiramon: Does the citation below require brackets?
% Solved
based \cite{chatgpt}. 
% The training procedure uses auto-regression where it learns to predict the  next token given a sequence of tokens. In order to perform this task correctly, it not only needs to learn the underlying language, but also, to some extent, commonsense knowledge around how language is used. This arms the model to
GPT-3 is able to perform competitively on several tasks such as question-answering, semantic parsing \cite{LLM_sem_parse} and machine translation. However, such LLMs tend to make simple mistakes in tasks such as semantic (commonsense) and mathematical reasoning \cite{gpt3-scope,chain,chatgpt-critic}. 

\subsection{Answer Set Programming and s(CASP)}
% Modify from the same part in the STAR paper

Answer Set Programming (ASP) is a logic programming paradigm suited for knowledge representation and reasoning \cite{cacm-asp}. ASP also facilitates commonsense reasoning \cite{gupta-csr}.  
The s(CASP) system, developed by \citealp{scasp}, is an answer set programming system that supports predicates, constraints over non-ground variables, uninterpreted functions, and, most importantly, a top-down, query-driven execution strategy.
These features make it possible to return answers with non-ground variables (possibly including constraints among them) and compute partial models by returning only the fragment of a stable model that is necessary to support the answer to a given query.
The s(CASP) system supports constructive negation based on a disequality constraint solver, and unlike Prolog's negation as failure and ASP's
default negation, %
{\tt not p(X)} can return bindings for \texttt{X} on success, i.e., bindings for which the call \texttt{p(X)} would have failed. 
%Additionally, s(CASP) system's interface with a constraint solver (over reals) allows for sound non-monotonic reasoning with constraints (useful for solving algebra problems in one of the NLU applications we discuss later).

Complex commonsense knowledge can be represented in ASP and the s(CASP) query-driven predicate ASP system can be used for querying it \cite{gelfondkahl,murder-trial}. Commonsense knowledge can be emulated using (i) default rules, (ii) integrity constraints, and (iii) multiple possible worlds~\cite{gelfondkahl,gupta-csr}. Default rules are used for jumping to a conclusion in the absence of exceptions, e.g., a bird normally flies, unless it's a penguin. 

%Thus, if we are told that Tweety is a bird, we jump to the conclusion that Tweety flies. Later, if we are told that Tweety is a penguin, we withdraw the conclusion that Tweety can fly. Default rules with exceptions represent an elaboration-tolerant way of representing knowledge~\cite{gelfondkahl}. 

\begin{lstlisting}[style=MySCASP]
flies(X) :- bird(X), not abnormal_bird(X).
abnormal_bird(X) :- penguin(X).
\end{lstlisting}

%{\small \tt
%flies(X) :- bird(X), not abnormal\_bird(X).
%
%abnormal\_bird(X) :- penguin(X).}  

\noindent Integrity constraints allow us to express impossible situations and invariants. For example, a person cannot sit and stand at the same time.

\begin{lstlisting}[style=MySCASP]
false :- person(X), sit(X), stand(X).
\end{lstlisting}

\noindent Finally, multiple possible worlds allow us to construct alternative universes that may have some of the parts common but other parts inconsistent. For example, the cartoon world of children's books has a lot in common with the real world (e.g., birds can fly in both worlds), yet in the former birds can talk like humans but in the latter, they cannot. 

%GG: paragraph below can be removed 
Default rules are used to model a bulk of our commonsense knowledge. Integrity constraints help in checking the consistency of the information extracted. Multiple possible worlds allow us to perform assumption-based reasoning (for example, knowing that ``Alice loves Bob", we could assume that either Bob also loves Alice or he does not). 

A large number of commonsense reasoning applications have already been developed using ASP and the s(CASP) system \cite{blawx,logical-english,chef,murder-trial}. %In the three applications reported in this paper, we have kept the  commonsense reasoning component simple, as our main goal is to illustrate our framework for combining LLMs and ASP to develop NLU applications that are explainable and reliable. 

The s(CASP) system is crucial for the AutoConcierge system. It holds the commonsense knowledge that a hotel concierge is expected to know about area restaurants, people's habits, cuisines, etc. With this commonsense knowledge, s(CASP) can detect inconsistencies in the text by reasoning over the predicates extracted. Justification for each response can also be given as the s(CASP) system can generate justifications as proof trees \cite{scasp-justification}.

%\subsection{STAR Framework}
%The STAR framework, that stands for ``Semantic-parsing Transformer and ASP Reasoner" \cite{star}, leverages both the advantage of LLMs and ASP systems by combining them together in a systematic way as discussed earlier. It parses the semantics of the text sentences and generates the predicates using LLMs such as GPT-3 and sends the predicates to the ASP system to get reliable answers through reasoning. Our AutoConcierge is designed based on the thought of the STAR.

\section{Design Philosophy}

%\subsection{Design of Modules}
% Separate reasoning part and LLMs
The main philosophy we follow in realizing a conversation agent is to emulate how humans process dialogs. Thus, AutoConcierge emulates a human concierge. When humans hear a sentence, they parse it to extract its meaning and represent the meaning in their minds as knowledge. Humans will check for consistency and correctness of this knowledge through the use of additional (commonsense) knowledge that also resides in their minds. Next, humans will process the knowledge to find gaps in it (and seek the missing information) or draw new conclusions from it. The knowledge representing the conclusion or the missing information is converted into a natural language and communicated. 

Thus, there are three phases in the process that humans follow: converting a sentence to knowledge, processing the knowledge to draw conclusions, and converting the conclusions into a response sentence. The conversational agent architecture on which AutoConcierge is based, similarly, has three phases (Figure \ref{fig:design}), that are realized as three separate modules---a module for predicate extraction, a module for reasoning, and a module for response generation. The reasoning module contains all the commonsense knowledge needed by AutoConcierge for generating a response (e.g., normally, curry dishes are part of Indian or Thai food). Our idea is to make the three parts self-contained while having them communicate with each other through clear, well-defined interfaces. Therefore, an intermediate filter is designed to parse and filter out irrelevant information from the output of the LLM before being passed to the reasoner. Following this structure, predicate generation is completely done by the Large Language Model GPT-3, all the reasoning is performed by the goal-directed s(CASP) system, while the natural language response generation is done by yet another invocation of GPT-3. 
%revisit the above again
%The mechanism of the AutoConcierge is illustrated in Figure \ref{fig:archi}.

\subsection{Translating Sentences to Predicates} 
\label{sec:s2p}

In our method, GPT-3 works as a ``translator" that translates sentences spoken by a human to predicates that can be understood by the machine. Essentially, we use GPT-3 purely as a semantic parser that generates predicates capturing the meaning (``deep structure") of the sentences. For simplicity, this set of predicates is designed by us and restricted to a specific domain.
%for translating sentences into predicates that capture the sentence's meaning.
This parsing step is realized by prompting the GPT-3 model with a few examples---referred to as in-context learning. An instance is shown below:

\begin{quote} There is a restaurant in the city center, Alimentum, which is not family-friendly. \#\#\# restaurant-name(alimentum), establishment(restaurant), family-friendly(no)
\end{quote} 

\noindent The example contains a sentence and the corresponding (essential) predicates we want to extract from its meaning, separated by `\#\#\#'. GPT-3 will extract these predicates even if the sentence is syntactically different (e.g., Alimentum is a family unfriendly restaurant in the city center). 
These examples are carefully selected so that they cover most of the cases, where each case expresses the concept in a different way. We also provide the type of predicates, as well as values when the scope of the values is fixed, for in-context learning. 
In-context learning can also be employed to identify the question that a user is interested in answering. For example, GPT-3 can be prompted to generate the predicate {\tt restaurant-name(query)} in response to the user's question ``Can you recommend a restaurant?" 
Some extra natural language instructions are also added at the beginning of the prompt to make GPT-3 better understand the task.

%GG: what about asking for history and another answer

\iffalse
\begin{table*}[h]
    \centering
    \begin{tabular}{c}
    \toprule
         Examples \\ \hline
         There is a restaurant in the city center, Alimentum, which is not family-friendly. \#\#\# name(alimentum), establishment(restaurant), family-friendly(no)
         \bottomrule 
    \end{tabular}
    \caption{Examples for the LLMs' Few-shot Learning}
    \label{tab:llm_exp}
\end{table*}

E2E is a dataset \cite{novikova2017e2e} containing restaurant information (in a form similar to predicates) and corresponding natural language descriptions. We employ in-context learning where we give the LLM examples of pairs containing a natural language sentence and the corresponding predicates.

The dataset contains the following eight predicates: \textbf{name}, \textbf{eatType} (as establishment type), \textbf{food} (as cuisine), \textbf{priceRange}, \textbf{customerRating}, \textbf{familyFriendly}, \textbf{area}, \textbf{near}. 
The \textbf{area} and \textbf{near} predicates are not included in the output during in-context learning, which is a cue to GPT-3 to ignore that information in the input sentence (later we add an address predicate to tell the user where the restaurant is located).
\fi 

What is remarkable is that a small number of such examples are enough for GPT-3 for in-context learning. This is due to GPT-3 being pre-trained on a vast amount of human-generated text \cite{gpt3}. Thus, 11 selected examples that cover all the predicates along with the possible arguments for each predicate suffice, including user queries, for AutoConcierge. 
%This ensures that the LLM is aware of every possible predicate as well as every possible argument value these predicates can take. 
We tested GPT-3 specialized to our in-context learning on the E2E dataset \cite{novikova2017e2e}, a dataset containing (i) restaurant information in a form similar to predicates and (ii) corresponding natural language descriptions. We adapted the first 500 examples in the E2E dataset for our specialized GPT-3 model (i.e., GPT-3 augmented with in-context learning) and obtained an accuracy of 89.33\%. The accuracy metric we use is designed to account for the proportion of predicates produced with their correct arguments. The high accuracy of predicate generation supports the feasibility of using the specialized GPT-3 as a semantic parser for AutoConcierge. 

\subsection{Commonsense Reasoning}

Once a sentence has been translated into predicates, the generated predicates are input into the reasoning module to compute the missing information. Or, if it determines that all the information has been acquired, it finds the restaurant to be recommended.

\smallskip 
\noindent{\bf Knowledgebase:}
The AutoConcierge system aims to recommend a user's local restaurants that satisfy a his/her preferences. To achieve this, information about local restaurants is necessary. For each restaurant, nine properties have been collected: the name of the restaurant, its cuisine type, its establishment type (bar, restaurant, coffee shop, etc.), its distance from the location of the concierge, its location address, its contact numbers, its price range, the average rating from the reviewers, and whether it is suitable for a family. This information is stored in the s(CASP) knowledgebase. 
Note that the predicate representation of these nine properties, plus additional predicates (i) indicating user preferences, (ii) that the user is interested in another answer, and (iii) the user is interested in a past answer, respectively, were used for GPT-3's in-context learning. The in-context learning we provided to GPT-3 also helps it identify if a statement relates to the user being thankful or saying pleasantries. If no predicate of interest is identified in a sentence, it is labeled as irrelevant. 
%GG: revisit; we should specify all the meaning predicates somewhere

\smallskip 
\noindent{\bf Conversational Knowledge Template:}
% Also mention the state and info-list design
In a conversation where we are trying to help someone with a specific task, we have a plan in mind regarding how to provide that help. The plan generally entails systematically asking for information from the other person, then reaching a decision. For each domain-specific task, the corresponding conversation plan has to be represented. We represent the plan as a \textit{conversational knowledge template} (CKT) that has been designed for implementing socialbots \cite{ckt}. 
%
%To make good recommendations for restaurants, AutoConcierge needs to know preferences of the user from the dialog. To acquire this knowledge, we design the next-topic generator based on the Conversational Knowledge Template (CKT). 
A CKT can be thought of as a state machine that allows AutoConcierge to systematically ask for information from a human user. It can be represented as a set of ASP rules coded in the s(CASP) system. 
%modeled as set of rules that create a related but not discussed topic based on the current conversation status. It allows the agent to jump out of the current subject of the talk actively, which helps reach the target of the conversation. 
%
In our AutoConcierge system, the CKT is used to explore users' preferences by directly asking them about their preferred type of food, their budget, etc. We define a set of properties that should be asked to get this user-preference information. These properties correspond 1-to-1 to the predicates representing meaning discussed earlier. For each response, the agent picks one of the properties from the set that has not been discussed before and generates a question on it. After collecting all the preference information in the set, 
% Abhiramon: below may be re-phrased as 'the agent will recommend a restaurant using this set.'
the agent will generate an appropriate restaurant recommendation.

\noindent{\bf Preference-based CKT Navigation:}
The CKT can be thought of as a representation of human commonsense knowledge for achieving the task at hand. The task may need additional commonsense knowledge that has to be explicitly represented as well. For instance, apart from the properties defined in the knowledge base, the users may employ some other related concepts that stand for a group of different food types. For example, when people say ``I prefer curry", our commonsense knowledge tells us that they are referring to the cuisines that provide food with curry. As a result, we automatically select Indian or Thai as the food type due to the preference for curry.  
%Although some restaurants providing other cuisines may also offer curry-based dishes, we only imply these two since we are unable to check the menu of each restaurant, and almost all the Indian, Thai, and Japanese restaurants provide food with curry. 
This strategy is also applied to other concepts. For example, if a person likes pizza, he or she might be seeking an Italian or American restaurant; and if someone is planning to drink alcohol, a pub or a bar is a good place to go. As much of this commonsense knowledge as possible---that we expect a human concierge to know---must be explicitly represented in the reasoning module. 

The predicates that are needed for decision-making are considered in the order in the next-topic selection step that is part of the CKT, implemented as an ASP program in the s(CASP) system. The order of these predicates is pre-defined in the reasoning module via a list, and the CKT always picks the property with a higher rank to generate the next question.
%, and thus this property should be satisfied at a higher level. 
However, if the users consider some properties more important than others, they can always ask for these requirements to be satisfied at the beginning of the 
% Abhiramon: can we say 'conversation' instead of 'talk' below
dialog. Likewise, a user can change their mind at any time, and the order coded in the CKT is overridden. 
%Yankai:
Thus, while searching for a restaurant recommendation, the reasoning module follows a default order defined by CKT, however, this order may be altered by user's preferences indicated in the conversation. We refer to the reasoning module that contains the CKT as the \textit{reasoner}.
% Thus, the CKT follows a default order in which information is acquired. This default order may be altered by a user's preferences as the dialog between the user and AutoConcierge occurs.

\subsection{AutoConcierge Response Generation}

Once a target restaurant is identified by the reasoner, its detailed information is presented in the form of logical predicates, containing the information that the user is seeking and the corresponding restaurant details. These predicates are translated into natural language sentences using GPT-3 in a reverse example format. For instance, the in-context learning example for this module is structured as follows:
%a few-shot learning example is structured as follows:

\begin{quote}
place(0, name, The Waterman), place(0, food type, Japanese), place(0, phone number, 414-247-2758) \#\#\# Perhaps you are interested in The Waterman, which offers Japanese cuisine. To make a reservation, you can call 414-247-2758.
\end{quote}

In the case of intermediate question generation, the reasoner outputs the property for the next question. A template is then applied to this property to generate a question, for example: \textit{Do you have any preference for the [property] of the place?}
Similarly, if the agent fails to find a satisfying restaurant, or the user is saying something irrelevant, the reply is generated based on the corresponding template.

To make the response more polite, natural, and varied, we further rephrase the generated reply using another GPT-3-based wording module, enhancing the agent's human-like quality with respect to producing a natural language response.

\begin{figure*}[ht]
    \centering
    \includegraphics[width=0.8\linewidth]{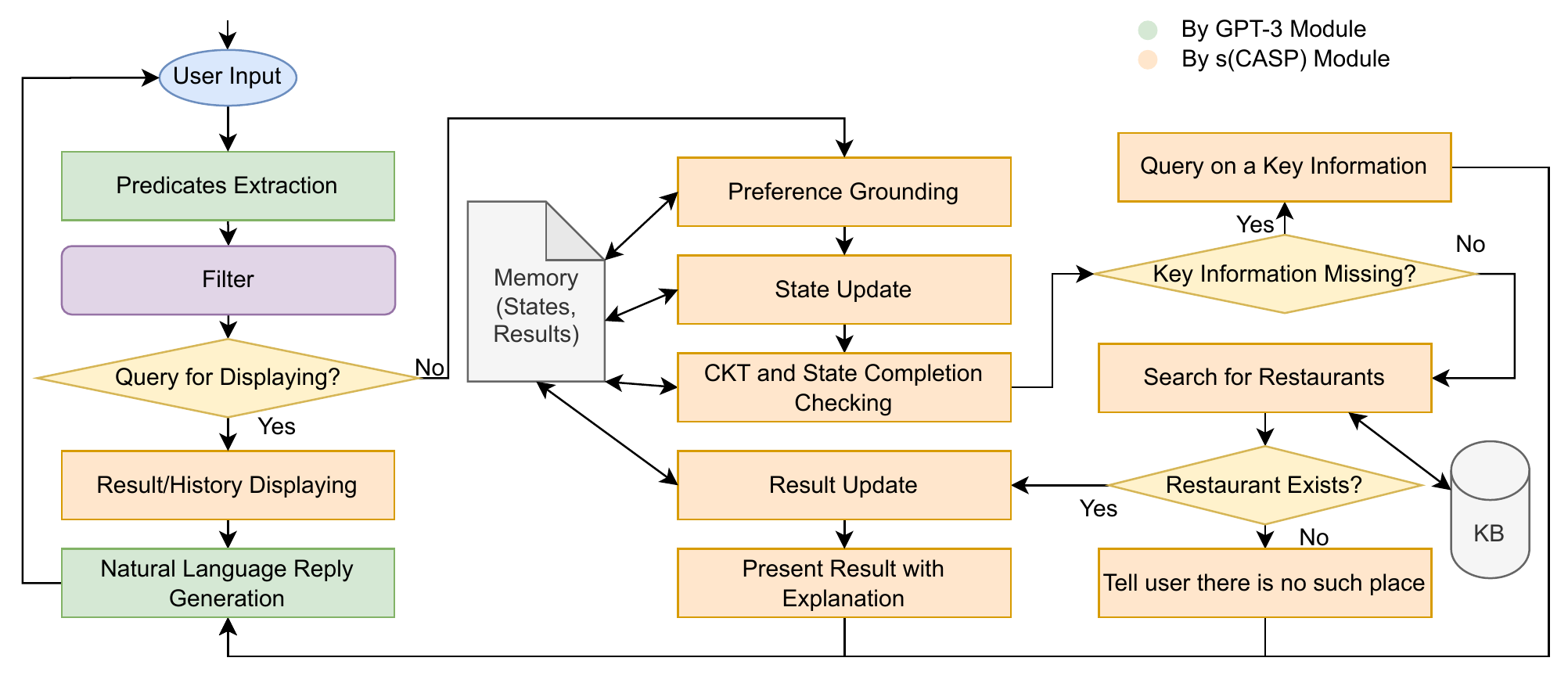}
    \caption{AutoConcierge Architecture; Green-colored boxes are handled by GPT-3, orange-colored by the s(CASP) system.}
    \label{fig:archi}
\vspace{-0.15in}
\end{figure*}

\section{AutoConcierge Implementation}
\label{sec:impl}

We provide specific details of the AutoConcierge system in this section. Fig. \ref{fig:archi} shows high-level details of how the system works. 

\subsection{Knowledge Representation}
The knowledge base collects nine properties for each restaurant: \textit{name}, \textit{food type}, \textit{establishment type}, \textit{price range}, \textit{customer rating}, \textit{address}, \textit{phone number}, \textit{family-friendliness}, and \textit{distance}.
%
% Abhiramon: 'Each' instead of 'one' below
Each property for a specific restaurant is expressed as a predicate in the following example format: 
$$\textbf{place}(\textit{restaurant\_ID}, \textit{property\_name}, \textit{property\_value})$$
For simplicity, the \textit{price range} predicate is defined for only three values: expensive, moderate, and cheap. Likewise, the \textit{customer rating} predicate is defined for three values: high, average, and low.
%
%are defined for three possible value . To make it not confusing to LLMs, we name the \textit{price range} as expensive, moderate, and cheap, while 
% Abhiramon: Can we say 'recording the customer rating as high, average, and low' Evaluation might mean something different.
%evaluating the \textit{customer rating} in high, average, and low. Similarly, the \textit{distance} consists of two values, near and far.
%
The \textit{food type} of a restaurant is usually its cuisine. When a restaurant's cuisine is not obvious, we record the food type it mainly serves. For example, the \textit{food type} of a fried chicken restaurant is chicken. The \textit{food type} of food shops and drink shops is marked as the food or drink they are serving. The \textit{food type} of different bars is set as ``bar", while the bar type can be further distinguished through the \textit{establishment} predicate that is defined for a regular restaurant, a shop, or a fast-food restaurant.
%The difference in the functions of the building is distinguished by \textit{establishment}. For example, the \textit{establishment} of a food place can be either a restaurant, shop, or fast-food. 
In the longer run, these concepts and associated commonsense knowledge have to be formally represented via a well-defined ontology, and constitute future work for us. 

\subsection{Interaction of Modules}
\label{sec:inter}
% Also include: Syntax Check
The AutoConcierge system consists of (i) GPT-3 specialized with in-context learning that serves as a semantic parser, (ii) a reasoner, and (iii) a response generator subsystem that converts predicates into text. The specialized GPT-3 model takes the natural language sentence as input and generates predicates that represent the sentence's meaning. The input is usually the user's reply, but when AutoConcierge asks for the user's (positive or negative) preference, the question posed by AutoConcierge is also made as part of the input to  GPT-3 augmented with in-context learning. The predicate domains are restricted to the nine properties in the knowledgebase, as well as extra ``prefer" and ``not\_prefer" that are used to capture user's additional preferences. The information the user is querying for (e.g, name of a restaurant, or its address) is indicated as ``query".
The format of the predicates generated by the predicate generation module is shown in the following example:

\begin{lstlisting}[style=MySCASP]
restaurant-name(query), 
price range(moderate),
establishment(restaurant, bar),
prefer(spicy, noodle),
address(query)
\end{lstlisting}

%\smallskip 
\noindent The predicates in the example indicate that we need to find a restaurant's name and its address. 
The in-context learning aid we provide to GPT-3 allows for predicates to contain multiple arguments. In  predicates other than ``prefer", this denotes the disjunction of these values, whereas in the ``prefer" or ``not\_prefer" predicates, it denotes a conjunction. This is designed for simplicity of the parsing phase, based on the characteristic of the properties. For properties such as \textit{food type}, \textit{customer rating}, etc., a restaurant can only have one value for these properties. However, a user may ask for ``Indian or Thai" food, for example, and both options chosen must be recorded. For the ``prefer" predicate, however, it is possible that one may seek a restaurant, for example, with both spicy food and noodles, and preference for both should be recorded. Likewise, for ``not\_prefer" predicate, one may indicate disliking Italian and Indian food. The options for preferences can be more complex, of course.

There are also two special predicates generated by the GPT-3 based parser, ``another\_option" and ``view\_history", which are used, respectively, to seek the next possible recommendation according to the current preference list and to keep track of the restaurants that the agent has recommended previously in the dialog.

In addition to the above predicates, our specialized GPT-3 model also classifies the input sentences into different labels, such as ``\textbf{irrelevant}" and ``\textbf{thank}". If a sentence is classified as ``irrelevant" by the parser, a filter that sits between the parser and the reasoner will generate a courteous response like ``Sorry, I am only a concierge helping with my users. Can I assist you with a restaurant recommendation?" The ``\textbf{thank}" label represents a sentence from the user that is expressing gratitude, and an appropriate response is generated (e.g., ``You are welcome"). The filter also works as a format modifier that adapts the parser output to the format the reasoner accepts. This is illustrated in Figure \ref{fig:archi}.

The format of the information provided as input to the reasoner is slightly different from the parser output. The extracted predicates in the above example are rewritten as terms to fit the format of the reasoner as shown below.
\smallskip 

\begin{lstlisting}[style=MySCASP, numbers=none, xleftmargin=0mm,  basewidth=.55em]
require('name', ['query']),
require('price range', ['moderate']),
require('establishment', ['restaurant','bar']),
require('prefer', ['spicy','noodle']),
not_require('food type', ['indian','thai']),
require('address', ['query'])
\end{lstlisting}

\smallskip 
\noindent The filter represents all the properties using just two predicates: ``require/2" and ``not\_require/2". This makes our implementation more flexible.

\subsection{Recommendation Generation}
\noindent{\bf CKT Implementation:} 
In AutoConcierge, at the top level, the CKT keeps asking questions of the user until all the information is collected. The properties \textit{food type}, \textit{price range}, and \textit{customer rating} are the key pieces of information that are acquired from the user. Users may state their preference positively (e.g., ``I want a cheap price range") or negatively (e.g,. ``Don't want it too expensive"). After each piece of information is acquired, the state is updated. Once all the information is available, the reasoner searches the querying property present in the state (e.g., 
{\tt restaurant-name(query)}). 
%
%While doing the recommendation, the properties the user is querying for are first analyzed and added to a querying list. 
If no querying property is found, a query for the \texttt{restaurant-name} is automatically added. The reasoner then finds all possible restaurants that satisfy the property preferences. %using the method introduced in Section \ref{sec:prefer}. 
The high-level code for the CKT is shown below. Note that {\tt neg\_member} relies on \textit{negation as failure}. The reasoner runs on the s(CASP) goal-directed ASP system \cite{scasp}. 

\begin{lstlisting}[style=MySCASP, basewidth=.52em]
next_action('ask',Question) :- state(State), 
   next_info(Question, State).
next_action('recommend', Result) :- 
   state(State), recommend(State, Result).

next_info(X,State) :- key_info(L), member(X,L), 
   member(require(X, ['query']), State).
next_info(X,State) :- key_info(L), member(X,L),   
   forall(A,neg_member(require(X, A), State)), 
   forall(B,neg_member(not_require(X,B),State)).

recommend(State, Result) :- 
   merge([require('name',['query'])],State,Next),
   recommend_(Next, Result).
recommend_(State, Recommendation) :-
   get_query_list(State, Queries),
   get_satisfied_places(State, X),
   fill_query(X, Queries, Recommendation).
\end{lstlisting}

The rule for {\tt get\_query\_list/2} predicate extracts properties that are queried, while the rules for {\tt get\_satisfied\_places/2} and  {\tt fill\_query/3} predicates generate the response, that is passed on to the next module that will generate a natural language response for the user. 

\iffalse
fill_query(X, [], []).
fill_query(X,[Attr|As],[place(X,Attr,Values)|Rs])
:- place(X, Attr, Values), fill_query(X, As, Rs).

get_query_list(State, []) :- 
  forall(Attr,neg_member(require(Attr,
                          ['query']),State)).
get_query_list(State, [Attr|R]) :- 
  select(require(Attr, ['query']), State, L1),
  get_query_list(L1, R).

get_satisfied_places(State, X) :- 
  select_requirement(State, State_Option),
  requirement_satisfy(X, State_Option).
\fi 

\smallskip 
%GG: this can be merged into previous section
\noindent{\bf Updating State:}
%
%After modification by the filter, the predicates obtained from a sentence are sent to the reasoner for computing a response. The response is in the form of a predicate, which is then converted into a natural language sentence by the third GPT-3 aided module. 
%
%If ``another\_option" and ``view\_history" are among these predicates are then used to update the current state. 
The CKT maintains a state which is represented by the list of ``require" and ``not\_require" structures that remember each selection the user makes during the conversation. The state is updated as the conversation progresses. The state is updated not only as a result of the user response, but also through the user's indication of (positive or negative) preferences. That is, the user may directly state that he/she likes Chinese food, for instance, or indicate a preference for noodles, in which case, liking for Chinese, Thai or Japanese food should be added to the state in the ``require" list. Users can also indicate a negative preference (``I do not wish to eat spicy food", for example). Non-preference for Thai and Indian food will be recorded in the ``not\_require" list in such a case. 

Once a negated property is recorded in the state, no matter whether the non-negated property exists, the reasoner would not ask for this property again. Meanwhile, when a state is updated with preference, any negative preference wrt this property will be removed from the state. In the recommendation step, these negative preferences act as constraints that any of the results cannot violate.

To update the state, we first extract the ``prefer" and ``not\_prefer" predicates from the dialog and find the corresponding \textit{food type} property. 
%The details are discussed later in  Section \ref{sec:prefer}. 
The newly acquired requirements are merged into agent's state. A priority order is maintained among the predicates that constitute the state. 
%GG: Ask Yankai for sentence below; done
%If the property to be added has a ``query" value captured in the state, the ``query" value is then covered by the updating value. 
If the state contained a predicate whose value was ``query"  (because the user had asked for the information), but new dialogs reveal a more specific value, then ``query" will be overwritten with that value. Similarly, if the user had indicated a negative-preference, but later changed his/her mind, the information will be updated as well. The {\tt update\_state/3} predicate below shows the high-level code.
%Note that users can indicate their dislike as well. These are recorded as part of the state using the {\tt not\_require} structure.
%If the value used to be set as dislike, when updating the state the ``dislike" predicate for this property value is removed. (For the ``not prefer" settings, please refer to Section \ref{sec:neg}.) The updated state list will not contain any ``prefer" property, and each of the other properties shows up at most once. The partial code for updating the state is shown below.

\begin{lstlisting}[style=MySCASP, basewidth=.52em]
update_state(L1, L2, L3) :- 
   set_prefer(L1, L4, Recm_List), 
   set_not_prefer(L4, Not_Recm_List, L5), 
   add_state(L5, L2, L6), 
   add_state(Recm_List, L6, L7), 
   add_state(Not_Recm_List, L7, L3).
\end{lstlisting}

\iffalse
add_state([], Updtd, Updtd).
add_state([require(Attr,Vals)|R],State,Updtd):-
   member('query', Vals), 
   member(require(Attr,Vals1), State),
   add_state(R, State, Updtd).
add_state([require(Attr,Vals)|R],State,Updtd):-
   neg_member('query', Vals), 
   member(require(Attr, Vals1), State), 
   select('query', Vals1, Vals2), 
   merge(Values2, Vals, Vals3), 
   replace(require(Attr, Vals1), 
           require(Attr, Vals3), State, Next), 
   remove_not_require(Next, Attr, Vals, Next1),
   add_state(Rest, Next1, Updtd).
add_state([require(Attr, Vals)|R],State,Updtd):-
   neg_member('query', Vals), 
   member(require(Attr, Vals1), State), 
   neg_member('query', Vals1), 
   merge(Values1, Vals, Vals2), 
   replace(require(Attr, Vals1), 
           require(Attr, Vals2), State, Next),
   remove_not_require(Next, Attr, Vals, Next1), 
   add_state(Rest, Next1, Updtd).
add_state([require(Attr,Val)|R],State,Updtd):-
   forall(V,neg_member(require(Attr,V),State)), 
   append(State,[require(Attr,Val)],Next), 
   remove_not_require(Next,Attr,Val,Next1), 
   add_state(R, Next1, Updtd).
\fi

\smallskip\noindent{\bf Computing Responses:} At each step, the reasoner examines the state, finds the information that it is still missing, crafts a term, and passes it to the response module that uses GPT-3 to translate it  into a natural language response or question.  Once all the user querying is done, the reasoner finds the restaurants that fit the criteria. An appropriate term is crafted again, and communicated to the response module. If there are no answers (because the criteria are too strict, for example) the representation of ``no answer found" is communicated to the response module. The response module generates the natural language response. It either gives the answer or asks the user to relax the criteria. If there are multiple restaurants that satisfy the criteria, they are stored in an output list. The first one in the output list is returned, and is subsequently removed from the output list to another list that maintains the history of recommendations. 
%Yankai edit
If the user rejects the current recommendation and asks for another one, the top one remaining in the output list is returned, and so on.
%However, if the user rejects the current recommendation and asks for another one, the next one in the output list is returned, and so on. 
%Every time the state is updated, the list of recommendation is set to an empty list. 
Note that our specialized GPT-3 model is trained to recognize sentences corresponding to the user asking for another restaurant (e.g., ``Can you recommend another one?"). In such a case, GPT-3 model extracts the predicate ``another\_option" as an input to the reasoner.

%The predicate ``another\_option" indicates to present the next result of recommendation other than the previously provided. Once the reasoner finds some results for a recommendation, the IDs of all possible restaurants are then stored in a result list. Each time the user is querying for a new solution, the top result is pop out, and the recommendation is given based on the information about this restaurant. Every update on the state will clean up the former recommendation list.

Similarly, the ``view\_history/1" predicate prompts the reasoner to recall previous recommendations. 
For instance, when the user asks ``Can you show me the restaurant you recommended at first?", 
%which is possible considering the occasion to convert the text to speech, 
GPT-3 will convey the semantics to the reasoner through the ``view\_history(first)" predicate.
%The reasoner maintains a pointer about which recommendation we are referring to, which is set by default to the recommendation just made after updating the state. 
The ``view\_history" predicate can take numbers as values indicating the $i^{th}$ previous recommendation, or ``first" or ``last" that refers to the first or the last recommendation. 
%Yankai: commented out the below line since it repeats the words
%Similar to the recommendation result, the history is recorded in a list for each dialog. 
The specialized GPT-3 model has also been trained to generate the ``view\_history" predicate for the appropriate sentence.

\subsection{Explanation}
One feature of AutoConcierge is that it is able to explain the reason for the generated recommendation. This is primarily because it has a complete understanding of user preferences and it uses commonsense reasoning to compute its response. Whenever a recommendation is made, the reasoner explicitly explains how the restaurant meets the user's criteria. If the agent is not able to make any recommendation based on the preference provided, it also explains and suggests a possible change in the preference that helps find a solution. 
% Yankai
Failure to recommend happens only if the search query fails. The reasoner can then run a negated query on s(CASP), that is guaranteed to succeed, then analyze the proof trace to find the property to relax. The property relaxed is the one that contributes to the success of the negated query (and thus was the cause of the original failed query). Details are omitted due to a lack of space. It should be noted that s(CASP) system's goal-driven nature and its ability to produce a justification \cite{scasp-justification} is crucial here.
%To achieve this, the reasoner traverses all the required properties, adding their highest priority value one after another to form a looser preference list, until the first search failure is reached. This helps the reasoner knows where the search fails, and it will list all other options in the property that fails to come out with a result, and suggest the user pick one. This allows the user to know where the search fails.

\section{Experiments and Results}

We have extensively tested the AutoConcierge system. We show an example dialog trace below (additional traces are included in the supplemental material). We show the state that the reasoner maintains after each user utterance for the conversation below. Note that because we use GPT-3, user dialogs don't have to be syntactically perfect, they can be slightly ungrammatical or colloquial.

\medskip  
%\noindent
%\textbf{Conversation 1}
\noindent \textbf{\large Conversation 1}
\begin{description}[align=right,labelindent=!,leftmargin=1cm]
  \addtolength{\itemsep}{-0.25\baselineskip}
  \tt \small
\item[Bot:] Hi there, how can I assist you?
\item[User:] Can you recommend me a restaurant?
\end{description}
\vspace{-0.1in}
\begin{lstlisting}[style=MySCASP, basewidth=.49em]
require('name',['query']),
require('establishment',['restaurant'])
\end{lstlisting}
\vspace{-0.1in}
\begin{description}[align=right,labelindent=!,leftmargin=1cm]
  \addtolength{\itemsep}{-0.25\baselineskip}
  \tt \small
\item[Bot:] Is there a particular type of food you would like to eat?
\item[User:] I can try any food except curry.
\end{description}
\vspace{-0.1in}
\begin{lstlisting}[style=MySCASP, basewidth=.49em]
require('name',['query']),
require('establishment',['restaurant']),
not_require('food type',['Indian','Thai'])
\end{lstlisting}
\vspace{-0.1in}
\begin{description}[align=right,labelindent=!,leftmargin=1cm]
  \addtolength{\itemsep}{-0.25\baselineskip}
  \tt \small
\item[Bot:] Are you looking for a certain price range of restaurants?
\item[User:]  Less than fifteen dollars.
\end{description}
\vspace{-0.1in}
\begin{lstlisting}[style=MySCASP, basewidth=.49em]
require('name',['query']),
require('establishment',['restaurant']),
not_require('food type',['Indian','Thai']),
require('price range',['cheap'])
\end{lstlisting}
\vspace{-0.1in}
\begin{description}[align=right,labelindent=!,leftmargin=1cm]
  \addtolength{\itemsep}{-0.25\baselineskip}
  \tt \small
\item[Bot:]  Are you looking for a place with a particular customer rating?
\item[User:]  No, I'm not looking for a specific rating score.
\end{description}
\vspace{-0.1in}
\begin{lstlisting}[style=MySCASP, basewidth=.49em]
require('name',['query']),
require('establishment',['restaurant']),
not_require('food type',['Indian','Thai']),
require('price range',['cheap']),
require('customer rating',['low','average','high'])
\end{lstlisting}
\vspace{-0.1in}
\begin{description}[align=right,labelindent=!,leftmargin=1cm]
  \addtolength{\itemsep}{-0.25\baselineskip}
  \tt \small 
\item[Bot:] Southern Recipes Grill has an average customer rating and serves budget-friendly American cuisine.
\item[User:]   Sounds nice. Can you give me its address?
\end{description}
\vspace{-0.1in}
\begin{lstlisting}[style=MySCASP, basewidth=.49em]
require('name',['query']),
require('establishment',['restaurant']),
not_require('food type', ['Indian','Thai']),
require('price range',['cheap']),
require('customer rating',['low','average','high']),
require('address',['query'])
\end{lstlisting}
\vspace{-0.1in}
\begin{description}[align=right,labelindent=!,leftmargin=1cm]
  \addtolength{\itemsep}{-0.25\baselineskip}
  \tt \small
\item[Bot:] Located at 621 W Plano Pkwy \#229, Plano, TX 75075,
  Southern Recipes Grill offers delicious food at an affordable
  price. Customers have given it an average rating.
\item[User:]  Thank you for your help.
\item[Bot:]  It's my pleasure to help. No need to thank me.
\end{description}

\smallskip

\iffalse
\noindent
\textbf{Conversation 2}

\begin{description}[align=right,labelindent=!,leftmargin=1cm]
  \addtolength{\itemsep}{-0.25\baselineskip}
  \tt
\item[Bot:]  How can I be of service?
\item[User:]  Do you know where can I find a place for drink?
\item[Bot:]  Do you have a budget in mind for the location?
\item[User:]  At low price, please.
\item[Bot:]  Do you have any preference regarding the customer reviews of the establishment?
\item[User:]  I'd prefer those with good reviews.
\item[Bot:]  If you need a great-value coffee spot that has positive feedback from customers, Eiland Coffee is your best bet.
\item[User:]  Sorry I don't drink coffee.
\item[Bot:]  Fukuro is a popular spot known for offering great bubble tea at an affordable cost with high ratings from customers.
\item[User:]  Maybe a bar suits me better.
\item[Bot:]  I would suggest the Northside Drafthouse \& Eatery; it's located at 3000 North Blvd suite 800, Richardson, TX 75080, it is affordable and has good ratings from customers.
\item[User:]  Sounds nice! Thanks!
\item[Bot:]  It's my pleasure to be of service.
\end{description}

\smallskip
\fi

\noindent Note that AutoConcierge is able to change the wording each time it asks a question, thanks to the use of GPT-3 for question generation. We show two additional conversations below. Conversation 2 shows the case where the user repeatedly rejects the suggestion made by the bot. Conversation 3 illustrates the situation where a restaurant cannot be found within user's constraints and the bot suggests which constraints to relax. 
\bigskip

\noindent \textbf{\Large Conversation 2}
\smallskip
\begin{description}[align=right,labelindent=!,leftmargin=1cm]
  \addtolength{\itemsep}{-0.25\baselineskip}
  \tt
\item[Bot:]  How can I be of service?
\item[User:]  Do you know where can I find a place to drink?
\item[Bot:]  Do you have a budget in mind for the location?
\item[User:]  At low price, please.
\item[Bot:]  Do you have any preference regarding the customer reviews of the establishment?
\item[User:]  I'd prefer those with good reviews.
\item[Bot:]  If you need a great-value coffee spot that has positive feedback from customers, Eiland Coffee is your best bet.
\item[User:]  Sorry I don't drink coffee.
\item[Bot:]  Fukuro is a popular spot known for offering great bubble tea at an affordable cost with high ratings from customers.
\item[User:]  Maybe a bar suits me better.
\item[Bot:]  I would suggest the Northside Drafthouse \& Eatery; it's located at 3000 North Blvd suite 800, Richardson, TX 75080, it is affordable and has good ratings from customers.
\item[User:]  Sounds nice! Thanks!
\item[Bot:]  It's my pleasure to be of service.
\end{description}

\bigskip

\noindent\textbf{\Large Conversation 3} 
\smallskip
\begin{description}[align=right,labelindent=!,leftmargin=1cm]
  \addtolength{\itemsep}{-0.25\baselineskip}
  \tt
\item[Bot:]
Hi, how can I be of help?

\item[User:] 
I'm looking for somewhere serving pizza. I want to have dinner with my family.

\item[Bot:]
What price range are you looking for?

\item[User: ]
Please make it as cheap as possible.

\item[Bot:]
Are you looking for a specific customer rating for the place?

\item[User:] 
Yes. I want the high rating ones.

\item[Bot:]
Sorry, we couldn't find any results for your specifications. We could find results that meet the conditions of food type being Italian, the establishment type being a restaurant, and requiring the location to be family friendly, but we were unable to find an option that fit within your budget criteria.

\item[User:] 
How about change the price to average?

\item[Bot:]
If you're looking for somewhere to eat that's rated highly and doesn't break the bank, Cappuccino Italian Bistro may be the perfect option for your family.

\item[User:] 
Any other recommendations?

\item[Bot:]
Palio's Pizza Cafe offers quality pizza at an affordable price with plenty of positive customer reviews. It is a great option for families as well.

\item[User:] 
That's great! May I have its address?

\item[Bot:]
Cappuccino Italian Bistro is an ideal choice for a family meal. It is located at 1310 W Campbell Rd Ste 135, Richardson, TX 75080 and offers Italian cuisine at a reasonable cost. Customer ratings for this establishment are notable.

\item[User:] 
No, not this one. I mean the address for Palio's Pizza Cafe.

\item[Bot:]
Palio's Pizza Cafe is a great choice if you are looking for a spot with a kid friendly atmosphere at an average price point. Located at 1469 W Campbell Road in Richardson, Texas 75080, they serve high-quality pizza and have great customer reviews.

\item[User:] 
Cool. Thanks.                       

\item[Bot:]
I would be happy to assist you.

\end{description}

\begin{table*}[ht]
    \centering
    \begin{tabular}{l@{\hskip 4em}c@{\hskip 4em}c@{\hskip 4em}c}
    \toprule
      & Next Question (1 Req.) & Next Question (3 Req.) & Recommendation (5 Req.) \\
      \midrule
      Find First  & 0.624 & 0.626 & 0.637 \\
      Find All & 0.633 & 0.638 & 1.384 \\[.5em]
      End-to-End & 1.571 & 1.672 & 4.602 \\\bottomrule
    \end{tabular}
    \caption{Run-Time (s) comparison in different modes. We
      experimented with the question generation mode with one and
      three requirement(s) in the current state, and with the
      recommendation mode with five requirements.}
    \label{tab:mode}
\end{table*}

\iffalse 
\begin{table*}[ht]
    \centering
    \begin{tabular}{l@{\hskip 4em}ccc@{\hskip 4em}ccc@{\hskip 4em}ccc}
    \toprule
     & \multicolumn{3}{c@{\hskip 4em}}{\textbf{1 Query}} &  \multicolumn{3}{c@{\hskip 4em}}{\textbf{2 Queries}}  &  \multicolumn{3}{c}{\textbf{3 Queries}} \\
      \# of Negations    &0 & 1 & 2   & 0 & 1  & 2  &0  & 1  & 2  \\
      \midrule
       Find First  & 0.637 & 0.875 & 0.974 & 0.673 & 1.101 & 1.202 & 0.705 & 1.397 & 1.517 \\
       Find All & 1.384 & 5.425 & 5.663 & 1.160 & 3.167 & 3.271 & 1.049 & 1.951 & 2.012 \\
       Find Top 3 & 0.679 & 1.052 & 1.174 & 0.676 & 1.090 & 1.233 & 0.680 & 1.123 & 1.268\\[.5em]
       Real-time & 4.602 & 6.276 & 6.245 & 4.624 & 4.227 & 5.911 & 5.803 & 4.550 & 4.475 \\\bottomrule
    \end{tabular}
    \caption{Run-Time (s) comparison. Note that it increases when the
      user is querying for more information about the recommended
      restaurant. The negated preferences (user's dislikings) also
      affect the time consumption. During the experiment, the user's
      preference is fixed.}
    \label{tab:main}
\end{table*}
\fi 

\subsection{Performance Evaluation}

%GG: reasoner and CKT are synonymous 
%\subsection{AutoConcierge Dialog Examples}
%We have tested our AutoConcierge system manually over different inputs. We next give a conversation sample to show how our agent works. The back-end state is also displayed after each response from the user. More dialog sessions are given in the Appendix. 

The AutoConcierge system is reasonably efficient. 
We conducted experiments to evaluate the execution efficiency of our AutoConcierge system, with a focus on
%GG: reasoner component only? 
% Yankai: Yes
time taken for \textit{next question generation} and \textit{computing final recommendation} in the reasoning phase. Specifically, we measured the time required for these tasks, as presented in Table \ref{tab:mode}, where we varied the number of requirements for the next question generation. We set the number of requirements to five for the recommendation mode. We also measured the time taken by the reasoner in finding the first (partial) answer set as well as in finding all the 
%Yankai: what does partial mean here
(partial) answer sets. The time taken by the reasoner was measured via the s(CASP) system's command-line interface. The End-to-End row in Table \ref{tab:mode} includes the time for predicate generation, reasoning, and natural language response generation combined. We conducted 10 runs of each experiment and calculated the average results. Our findings, presented in Table \ref{tab:mode}, reveal that the number of requirements in the question generation stage does not significantly affect performance. Our results also show that GPT-3 components (parsing and response generation) take more time than the reasoning component. Note that these response times are quite fast and do not appear too slow to a human.
%, while generating recommendations 
%for all possible answers 
%requires more time than generating a single answer.

%Table \ref{tab:main} presents the complete experiments with the recommendation component, incorporating two additional factors: the number of information types that the user is querying and the number of negations in the requirements. Our findings indicate that an increase in the number of negations and queries results in slightly longer execution times, except for the case where all possible solutions are sought, in which the time consumption reduces with the number of queries increases. We attribute this exception to the pruning of the search field, which becomes more specific with additional queries. The experiments also reveal that real-time response times are more variable, as the parser relies on the GPT-3 model and requires responses over the internet. However, a response delay of 4 to 6 seconds is generally acceptable.

\subsection{Comparison with Bing AI}

We conducted an experiment to test restaurant recommendations using Microsoft's Bing AI \cite{bing}, a recently released GPT-based chatbot with a search engine. To ensure fairness, we limited the scope of the search for a restaurant to two miles around the University. The Bing AI integrates GPT-based sentence generation and search engine knowledge acquisition, and provides the source link for the information. The conversation from one experiment is presented below (dialogs are  abridged due to lack of space) 

\smallskip
\begin{description}[align=right,labelindent=!,leftmargin=1cm]
  \addtolength{\itemsep}{-0.25\baselineskip}
  \tt \small
\item[Bing:]  What can I assist you with now?
\item[User:]  Hi I'm now at UT Dallas, could you recommend me a nearby restaurant within 2 mile ride?
\item[Bing:]  Sure! There are several restaurants near UT Dallas within 2 miles. 
%Some options include The Pub at UTD which is located on the 2nd floor of Student Union. Another option is Aloha Chicken And Shrimp, a family-owned \& operated restaurant that offers delivery and takeout. 
(gives two top-result recommendations)
Is there a specific type of cuisine you’re interested in?
\item[User:]  I'm looking for Italian food.
\item[Bing:]  There are several Italian restaurants near UT Dallas within 2 miles. 
%One option is Nonna, located at 4115 Lomo Alto Dr. Another option is Saint Rocco’s, an authentic Little Italy-style restaurant. 
(gives another two examples for Italian restaurants)
Would you like more information on these restaurants?
\item[User:]  I don't have a high budget, are these restaurants expensive?
%\item[Bing:]  I’m sorry but I don’t have information on the prices of these restaurants. Would you like me to search for more affordable Italian restaurants near UT Dallas?
%\item[User:]  Yes please.
\item[Bing:]  One affordable Italian restaurant near UT Dallas within 2 miles is 
Jimmy’s Food Store, a specialty food market that offers exquisite Italian food. Another option is Botolino Gelato Artigianale, which serves Italian gelato. 
Would you like more information on these restaurants?
\item[User:]  Can you tell me where is Jimmy's Food Store?
%\item[Bing:]  I’m sorry but I don’t have the exact address for Jimmy’s Food Store. Would you like me to search for it?
%\item[User:]  Yes please.
\item[Bing:]  Jimmy’s Food Store is located at 4901 Bryan Street, Dallas. Is there anything else you would like to know?
\item[User:]  But it seems very far away from UT Dallas. Any other options?
\item[Bing:]  Yes, another option for affordable Italian food near UT Dallas within 2 miles is Botolino Gelato Artigianale, which serves Italian gelato. Would you like more information on this restaurant?
\item[User:]  Tell me more about Botolino Gelato Artigianale.
\item[Bing:]  
%Botolino Gelato Artigianale is a gelato shop located at 2116 Greenville Ave. They offer a variety of gelato flavors and frozen desserts made from scratch using traditional methods and pure ingredients. They also serve mascarpone, rolls, sandwiches and key lime pie. 
(gives details for Botolino Gelato Artigianale)
Would you like more information on this restaurant?
\item[User:]  No thanks. It is also very far away from UT Dallas. Please recommend the restaurant nearby within 2 miles drive.
\item[Bing:]  I’m sorry but I couldn’t find any affordable Italian restaurants within 2 miles of UT Dallas. Would you like me to look for other options?
\item[User:]  ...
%Okay. How about food with curry?
%\item[Bing:]  One option for food with curry near UT Dallas within 2 miles is Thai Spice located at 18111 Dallas Pkwy. They serve a variety of Thai dishes and are open for lunch and dinner. Would you like more information on this restaurant?
  \end{description}

\smallskip

As evident from the above Bing AI conversation, the GPT-based search engine, despite generating natural language responses similar to a human concierge, exhibits several deficiencies. These include:
\begin{itemize}
\item Bing AI does not ask for detailed user preferences like a human concierge would. It arbitrarily recommends restaurants. Any request for preference by Bing AI, e.g., asking for a specific type of cuisine, is driven by what the user said. In contrast, AutoConcierge actively seeks information to achieve the task goal.
\item Bing AI only recommends popular restaurants on the web. It cannot recommend restaurants that are not listed on any website. On the other hand, our AutoConcierge possesses knowledge of all restaurants and can make precise recommendations. It can even recommend a nearby grocery store to buy a sandwich if all restaurants nearby are closed. This is an obvious advantage that task-oriented bots have over generic search-engine-based systems such as Bing AI.
\item Bing AI is limited to generating only ten responses, which requires the user to provide accurate and detailed information in a few questions. It may sometimes ignore specific requirements, necessitating their repeated mention. The limitation to 10 responses is there because of Microsoft's worry that Bing AI will go haywire or can become offensive, etc. AutoConcierge, on the other hand, allows users to reply as many times as they like, and the agent remembers the requirements once it is mentioned. 
\item Bing AI can accurately convey information available through search, but it may hallucinate information if it cannot be found on the current page. For instance, in the example above, several of the restaurants recommended were more than 10 miles away, while the bot asserted that they were within 2 miles. AutoConcierge ensures that all information provided by the agent is reliable.
\end{itemize}

\subsection{Evaluation Principles}
Most of the current evaluation metrics in text generation and conversational tasks are based on the similarity comparison to the given expected answer. However, these metrics are only focused on single-reply conversations. For a conversational agent, where human interaction is involved along with proper natural language dialogs, these methods are deficient.
%since in each run, users can change their purpose, and the quality of the response generated by the agent is able to be improved during the conversation. The experiment also requires involving another ``smart" agent or human to interact in the conversation.
We could not find a metric for evaluating goal-oriented conversational agents, so to evaluate the performance of our AutoConcierge system, we propose six criteria for evaluating task-driven conversational agents.

\smallskip 
\noindent \textbf{Proactivity:} refers to how actively is the agent involved in the conversation. The agent should be focused on the task and not discuss off-topic things while keeping the conversation engaging.
%The goal of the conversation is to complete a specific task, so it is necessary for the agent to focus on the task and not discuss off-topic things while keeping the conversation engaging.

\smallskip 
\noindent \textbf{Economy:} indicates how quickly the agent finds a satisfactory result using a minimum number of exchanges. 
%A good conversational agent should be able to solve a user's problem with a minimum number of dialogs. 
Users should not have to repeat their questions multiple times to let the agent understand.

\smallskip 
\noindent \textbf{Explainability:} the agent should be able to explain each of its answers. 
%In a task-driven dialog, explainability is important for the user to understand the current situation and take the next action.

\smallskip 
\noindent \textbf{Correctness:} the agent should provide correct responses for the task at hand. It should ask the right questions to acquire knowledge from the user. 
%should be considered an essential metric, not only because it allows the information to be conveyed securely. It also helps the users not get confused in the conversation and enhances efficiency.

\smallskip 
\noindent \textbf{Consistency:} refers to the ability of the agent to remember all the requirements and satisfy as many as possible.
%An ideal task-driven conversational agent should make the best recommendation based on the knowledge it acquires.

\smallskip 
\noindent \textbf{Efficiency:} the correct answer should be computed within a reasonable time (\textit{bounded rationality}), else the user will lose interest.
%A reply that takes a very long time to generate can also kill the patience of the users.

%These evaluation metrics should be used in user-studies. However, we were 

With respect to the above metrics, AutoConcierge outperforms the state-of-the-art Bing AI. Bing AI does give an explanation but for recommendations that were irrelevant (restaurants were too far away despite exhortation to be within 2 miles). Efficiency was similar in both, as responses in both were generated within average human attentions span. In general, Bing AI's responses are heavily influenced by the question asked, unlike AutoConcierge. Bing AI's behavior is not surprising given that it is purely based on pattern matching with zero logical reasoning involved. Note that AutoConcierge is also safe, in that if the user uses any objectionable words in a sentence, it will be classified as irrelevant, and ignored, and a generic response issued.

%\section{Discussion}

%\subsection{Awareness}

%\subsection{Developing Bots for Other Domains}

%In this paper, we reported  a conversational agent in the restaurant recommendation domain. It excitedly reaches the target where the reply is generated consciously with a specific purpose and knowledge. This method can be easily applied to other tasks as long as they require collecting information from the user first and providing service according to this information. In the parser part, a few new examples for the target task are enough. The details of the reasoner may be different, but the core functions, the CKT and the %Preference Mapping
%CKT Navigation, can be used directly. Creating a template that generates the s(CASP) codes for different task-driven conversations is also feasible.

%Apart from the task-driven conversation, our method can also be applied to the gossiping conversation. With several topics given, the CKT allows the agent to jump among the topics freely, and 
%Preference Mapping 
%Preference-based CKT Navigation makes it possible to map different concepts to the structured knowledge in the knowledge base. The conversational agents with our method are more actively interacting with the user, and the information they provide is reliable.

\section{Related Work}

Conversational agents (chatbots) have been an active area of research for a long time. Many chatbots have been deployed by businesses, especially e-commerce websites. Most chatbots are based on machine learning technology. Some could be as simple as matching a user utterance to an answer in the FAQ (frequently asked questions) list. Most chatbots deployed in business are really deficient, and quickly handover to a human operator after one or two rounds of dialogs. Chatbots such as Amazon Alexa, Siri, and Google Assistant have also been developed, however, these are more of information retrieval devices rather than actual effective chatbots that can have a stateful conversation. The authors have participated in the Amazon Alexa Socialbot competition \cite{ckt}; much of the research reported in this paper is inspired by that effort. Amazon has invested heavily in developing task-bots \cite{alexa-taskbot} and social-bots \cite{alexa-socialbot} through University competitions. The bots developed as part of these competitions are still quite deficient. They rely heavily on machine learning technology, so do not understand what a user is saying. 

A recent line of related research uses different sizes of transformer-based models to build chatbots \cite{transfertransfo,dialogpt,meena,lambda,blenderbot3}. The number of parameters used in these models range from $\sim$117 Million to more than 175 Billion. These models 
are similar to GPT-3 \cite{gpt3}, and are based purely on pattern matching, and thus are similar to Bing AI and therefore suffer from similar disadvantages discussed earlier. 

%A recent line of related research uses different sizes of transformer-based models to build chatbots. Wolf et al. use a model of $\sim$117 M parameters that is fine-tuned on multi-turn conversation data \cite{transfertransfo}. They also use a next-sentence prediction loss along with language modeling loss that improves the \textit{perplexity} of the model. Zhang et al. train a transformer decoder-only model of $\sim$ 762 M parameters on exchanges in the comment section of Reddit ~\cite{dialogpt}. Meena \cite{meena} is a $\sim$2.6 Billion parameter transformer architecture model that is trained on human multi-turn conversation and has lower perplexity than the models mentioned above. They argue this leads to better sensibleness and specificity of conversation. Thoppilan et al ~\cite{lambda} train a $\sim$137 B parameter model on dialog data and web text. BlenderBot3 by ~\cite{blenderbot3}, is a $\sim$ 175 B transformer architecture model that is also designed to retrieve information from the internet when necessary to answer open-domain questions. Different from these approaches, our system does not exclusively rely on transformer-based models. While we use LLMs for English to predicate conversion, we use ASP for the reasoning component which makes our method more reliable.

Another aspect to consider for chatbots is how they are evaluated. \textit{Perplexity} is a popular automated metric that is used to measure chatbot performance ~\cite{transfertransfo}. It captures how `surprised' the model is during test time. Hence, better-performing models will have lower perplexity, which tends to the larger pre-trained models ~\cite{meena}. Other works use different ways of collecting human feedback. Shuster et al. allow users to give a thumbs down for a conversation and select if the bot was rude, off-topic, incorrect, etc. \cite{blenderbot3}. %Li et al. suggest that evaluators should choose one of the two speakers in the conversation and support their choice with reasons such as more engaging, more interesting, more knowledgeable, etc.  ~\cite{acuteeval}. 
XiaoIce \cite{xiaoice} is available online and is used by a large pool of users. They evaluate their model using conversation-turns per session arguing that an engaging chatbot will motivate users to interact longer with it. They also design their policy for selecting conversation-related actions to maximize this metric. Since we aim to assess the impact of using ASP for reasoning, we design metrics around reliability and explainability and compare the conversations of our model to similar conversations with a currently prominent model (Bing AI).
% Abhiramon: We may be able to compute perplexity of our model, but we can only compare it with other LLMs if we have access to the probabilities they produce for each token generated. (The API may have the option to return probabilities).
% Note also that perplexity measure for AutoConcierge is zero (i.e., there is no surprise for users in responses generated by AutoConcierge).  

Finally, AutoConcierge is an extension of our previous work developing NLU systems based on commonsense reasoning ~\cite{aaai21-kinjal,ckt}.  The use of GPT-3 with in-context learning as a semantic parser leads to a significant advantage over our previous work.  

\section{Conclusion and Future Work}
In this paper, we developed a domain-specific conversational agent called AutoConcierge for restaurant recommendations. AutoConcierge can understand human dialogs. It also communicates with the user through natural language dialogs. It interactively seeks information from a human user and converses in full sentences. AutoConcierge uses GPT-3 with in-context learning as a semantic parser to generate knowledge represented as predicates from sentences. This knowledge is processed further with the s(CASP) goal-directed ASP system that performs commonsense reasoning. Responses computed by s(CASP) are translated into natural language by leveraging GPT-3 again. The reasoner uses a  conversational knowledge template to organize the conversation and determine which questions to ask the user while taking user preferences into account. AutoConcierge outperforms Bing AI on the restaurant recommendation task. Other domain-specific task-bots, similar to AutoConcierge, can be readily developed for other tasks. Thus, AutoConcierge technology can be used to develop automated Bank Tellers, front desk receptionists in an office, socialbots \cite{caspr,alexa-socialbot}, etc. 
%a user-friendly front end for a complex software system for receiving user input,  

Our future work includes making the knowledge representation more general through the use of existing ontologies such as WordNet \cite{wordnet}. We also plan to develop a framework based on LLMs (GPT-3) and ASP (s(CASP)) for making the development of chatbots such as AutoConcierge significantly easier and faster. We also plan to develop a socialbot similar to AutoConcierge that can hold a conversation with humans about movies, books, sports, video-games, etc., similar to Amazon Alexa socialbots \cite{ckt,alexa-socialbot}. 

%We will continue to improve this conversational agent, adding more functions to the reasoner (e.g. knowledgebase ontology design, open hours checking), and making it more intelligent in replying. Another work to do is the generalizable template that applies this method to other similar tasks.

\section*{Acknowledgments}

Authors are partially supported by NSF grants IIS 1910131, IIP 1916206, US DoD, and several industry projects. We are grateful to the UT Dallas ALPS Lab members for discussions.

%% The file kr.bst is a bibliography style file for BibTeX 0.99c
\bibliographystyle{kr}
\bibliography{kr-sample}

\end{document}